\documentclass[11pt]{article}
\usepackage{emnlp2016}
\usepackage{times}
\usepackage{url}
\usepackage{latexsym}
\usepackage{xspace}
\usepackage{booktabs}

\usepackage{graphicx}
\usepackage{amsfonts}
\usepackage{microtype}
\usepackage{multirow}
\usepackage{url}
\usepackage{verbatim}
\usepackage{amsmath,amsthm,amssymb}
\usepackage{array}
\usepackage{ifthen}
\usepackage{courier}
\emnlpfinalcopy

\renewcommand{\b}[1]{\mathbf{#1}}

\usepackage{amsfonts}
\usepackage{microtype}
\usepackage[scaled=0.86]{helvet}

\newcommand{\captionfonts}{\small}
\makeatletter  
\long\def\@makecaption#1#2{%
  \vskip\abovecaptionskip
  \sbox\@tempboxa{{\captionfonts #1: #2}}%
  \ifdim \wd\@tempboxa >\hsize
    {\captionfonts #1: #2\par}
  \else
    \hbox to\hsize{\hfil\box\@tempboxa\hfil}%
  \fi
  \vskip\belowcaptionskip}
\makeatother   

\belowdisplayskip \abovedisplayskip

\usepackage{latexsym}

\author{Alessandro Sordoni$^f$ \and Philip Bachman$^f$ \and Adam Trischler$^f$ \and Yoshua Bengio$^g$ \\
$^f$Maluuba Research, Montr\'eal, Qu\'ebec \\
$^g$University of Montr\'eal, Montr\'eal, Qu\'ebec \\
\small \tt \{alessandro.sordoni, phil.bachman, adam.trischler\}@maluuba.com \\
\small \tt bengioy@iro.umontreal.ca}

\title{Iterative Alternating Neural Attention for Machine Reading}
\begin{document}

\maketitle
\begin{abstract}
We propose a novel neural attention architecture to tackle machine comprehension tasks, such as answering Cloze-style queries with respect to a document. Unlike previous models, we do not collapse the query into a single vector, instead we deploy an iterative alternating attention mechanism that allows a fine-grained exploration of both the query and the document. Our model outperforms state-of-the-art baselines in standard machine comprehension benchmarks such as CNN news articles and the Children's Book Test (CBT) dataset.

\end{abstract}

\section{Introduction}

Recently, the idea of training machine comprehension models that can read, understand, and answer questions about a text has come closer to reality principally through two factors.  The first is the advent of deep learning techniques~\cite{bengio_book}, which allow manipulation of natural language beyond its surface forms and generalize beyond relatively small amounts of labeled data.  The second factor is the formulation of standard machine comprehension benchmarks based on Cloze-style queries~\cite{hill2015goldilocks,hermann2015teaching}, which permit fast integration loops between model conception and experimental evaluation.

Cloze-style queries~\cite{taylor1953} are created by deleting a particular word in a natural-language statement. The task is to guess which word was deleted. In a pragmatic approach, recent work~\cite{hill2015goldilocks} formed such questions by extracting a sentence from a larger document. In contrast to considering a stand-alone statement, the system is now required to handle a larger amount of information that may possibly influence the prediction of the missing word. Such contextual dependencies may also be injected by removing a word from a short human-crafted summary of a larger body of text. The abstractive nature of the summary is likely to demand a higher level of comprehension of the original text~\cite{hermann2015teaching}. In both cases, the machine comprehension system is presented with an ablated query and the document to which the original query refers. The missing word is assumed to appear in the document.

Encouraged by the recent success of deep learning attention architectures~\cite{bahdanau2014neural,sukhbaatar2015end}, we propose a novel neural attention-based inference model designed to perform machine reading comprehension tasks.
The model first reads the document and the query using a recurrent neural network~\cite{bengio_book}.
Then, it deploys an iterative inference process to uncover the inferential links that exist between the missing query word, the query, and the document. This phase involves a novel alternating attention mechanism; it first attends to some parts of the query, then finds their corresponding matches by attending to the document. 
The result of this alternating search is fed back into the iterative inference process to seed the next search step. This permits our model to reason about different parts of the query in a sequential way, based on the information that has been gathered previously from the document. After a fixed number of iterations, the model uses a summary of its inference process to predict the answer.

This paper makes the following contributions. We present a novel iterative, alternating attention mechanism that, unlike existing models~\cite{hill2015goldilocks,watson}, does not compress the query to a single representation, but instead alternates its attention between the query and the document to obtain a fine-grained query representation within a fixed computation time. Our architecture tightly integrates previous ideas related to bidirectional readers~\cite{watson} and iterative attention processes~\cite{hill2015goldilocks,sukhbaatar2015end}. It obtains state-of-the-art results on two machine comprehension datasets and shows promise for application to a broad range of natural language processing tasks.

\section{Task Description}
\label{sec:task}
One of the advantages of using Cloze-style questions to evaluate machine comprehension systems is that a sufficient amount of training and test data can be obtained without human intervention. The CBT~\cite{hill2015goldilocks} and CNN~\cite{hermann2015teaching} corpora are two such datasets.

The CBT\footnote{Available at \url{http://www.thespermwhale.com/jaseweston/babi/CBTest.tgz}.} corpus was generated from well-known children's books available through Project Gutenberg. Documents consist of 20-sentence excerpts from these books. The related query is formed from an excerpt's 21st sentence by replacing a single word with an anonymous placeholder token. The dataset is divided into four subsets depending on the type of the word replaced.
The subsets are named entity, common noun, verb, and preposition. We will focus our evaluation solely on the first two subsets,~i.e. CBT-NE (named entity) and CBT-CN (common nouns), since the latter two are relatively simple as demonstrated by~\cite{hill2015goldilocks}.

The CNN\footnote{Available at \url{https://github.com/deepmind/rc-data}.} corpus was generated from news articles available through the CNN website. The documents are given by the full articles themselves, which are accompanied by short, bullet-point summary statements. Instead of extracting a query from the articles themselves, the authors replace a named entity within each article summary with an anonymous placeholder token.
\begin{table}[t]
\centering
\vspace{3mm}
\begin{tabular}{lrrrr}
\toprule
&  & CBT-NE & CBT-CN & CNN \\
\midrule
\# Train &  & 108,719 & 120,769 & 380,298      \\
\# Valid &  & 2,000    & 2,000      & 3,924    \\
\# Test  &  & 2,500    & 2,500      & 3,198    \\
\# Cand. ($|\mathcal{A}|$) & & 10 & 10        & $\sim$26 \\
Avg. $|\mathcal{D}|$ && $\sim$430 & $\sim$460 & $\sim$750 \\
\bottomrule
\end{tabular}
\caption{\label{tab:stats}Statistics of CBT-NE, CBT-CN and CNN.}
\end{table}

For both datasets, the training and evaluation data consist of tuples $(\mathcal{Q}, \mathcal{D}, \mathcal{A}, a)$, where $\mathcal{Q}$ is the query (represented as a sequence of words), $\mathcal{D}$ is the document, $\mathcal{A}$ is the set of possible answers, and $a \in \mathcal{A}$ is the correct answer. All words come from a vocabulary $V$, and, by construction, $\mathcal{A} \subset \mathcal{D}$. For each query, a placeholder token is substituted for the real answer $a$. Statistics on the datasets are reported in Table~\ref{tab:stats}.

\section{Alternating Iterative Attention}

\begin{figure*}[t]
    \centering
    \includegraphics[scale=1.2]{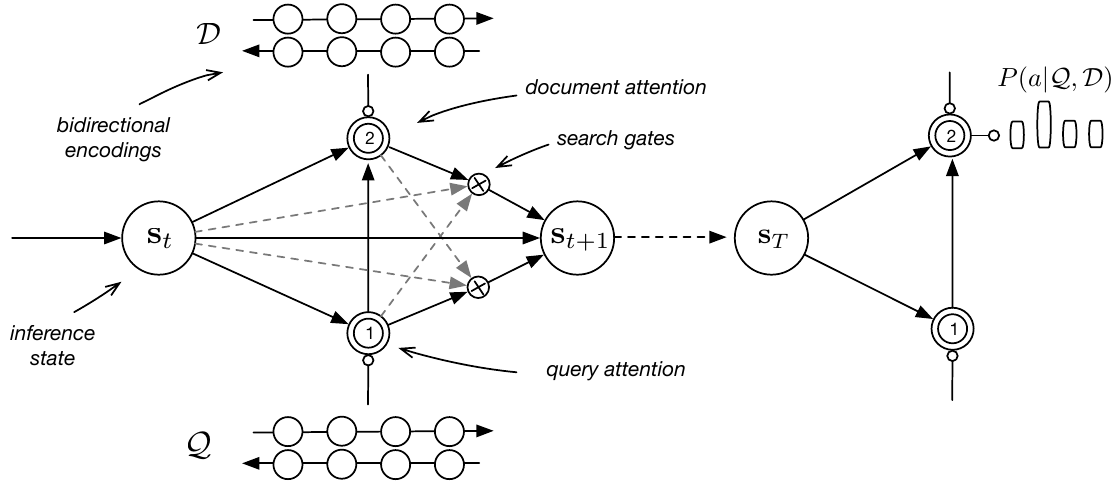}
    \caption{Our model first encodes the query and the document by means of bidirectional GRU networks. Then, it deploys an iterative inference mechanism that alternates between attending query encodings (1) and document encodings (2) given the query attended state. The results of the alternating attention is gated and fed back into the inference GRU. Even if the encodings are computed only once, the query representation is dynamic and changes throughout the inference process. After a fixed number of steps $T$, the weights of the document attention are used to estimate the probability of the answer $P(a|\mathcal{Q}, \mathcal{D})$.}
    \label{fig:model}
\end{figure*}

Our model is represented in Fig.~\ref{fig:model}. Its workflow has three steps.
First is the \emph{encoding} phase, in which we compute a set of vector representations, acting as a memory of the content of the input document and query.
Next, the \emph{inference} phase aims to untangle the complex semantic relationships linking the document and the query in order to provide sufficiently strong evidence for the answer prediction to be successful.
To accomplish this, we use an iterative process that, at each iteration, alternates attentive memory accesses to the query and the document.
Finally, the \emph{prediction} phase uses the information gathered from the repeated attentions through the query and the document to maximize the probability of the correct answer.
We describe each of the phases in the following sections.


\subsection{Bidirectional Encoding}
The input to the encoding phase is a sequence of words $\mathcal{X} =  (x_1, \ldots, x_{|\mathcal{X}|})$, such as a document or a query, drawn from a vocabulary $V$.
Each word is represented by a continuous word embedding $\mathbf{x} \in \mathbb{R}^d$ stored in a word embedding matrix $\mathbf{X} \in \mathbb{R}^{|V| \times d}$.
The sequence $\mathcal{X}$ is processed using a recurrent neural network encoder~\cite{bengio_book} with gated recurrent units (GRU)~\cite{GRU}. For each position $i$ in the input sequence, the GRU takes as input the word embedding $\mathbf{x}_i$ and updates a hidden state $\mathbf{h}_{i-1}$ to $\mathbf{h}_i = f(\mathbf{x}_i, \mathbf{h}_{i-1})$, where $f$ is defined by:
\begin{equation}
\label{eq:gru}
\begin{split}
& \b r_{i} = \sigma ( \b I_r\, \b x_i + \b H_r\b  h_{i-1}),\\
& \b u_{i} = \sigma ( \b I_u\, \b x_i + \b H_u\b  h_{i-1} ), \\
& \b{\bar h}_i = \text{tanh} (\b I_h\, \b x_i +\b  H_h (\b r_i \cdot \b h_{i-1})), \\
& \b h_i = (1 - \b u_i) \cdot \b h_{i-1} + \b u_i \cdot \b{\bar h}_i , \\
\end{split}
\end{equation}
where $\b h_i, \b r_i$ and $\b u_i \in \mathbb{R}^h$ are the recurrent state, the reset gate and update gate respectively, $\b I_{\{r,u,h\}} \in \mathbb{R}^{h \times d}$, $\b H_{\{r,u,h\}} \in \mathbb{R}^{h \times h}$ are the parameters of the GRU, $\sigma$ is the sigmoid function and $\cdot$ is the element-wise multiplication.
The hidden state $\mathbf{h}_i$ acts as a representation of the word $x_i$ in the context of the preceding sequence inputs $x_{<i}$. In order to incorporate information from the future tokens $x_{>i}$, we choose to process the sequence in reverse with an additional GRU~\cite{watson}.
Therefore, the encoding phase maps each token $x_i$ to a contextual representation given by the concatenation of the forward and backward GRU hidden states $\mathbf{\tilde{x}}_i = [\overrightarrow{\mathbf{h}_i}, \overleftarrow{\mathbf{h}_i}]$. We denote by $\mathbf{\tilde{q}}_i \in \mathbb{R}^{2h}$ and $\mathbf{\tilde{d}}_i \in \mathbb{R}^{2h}$ the contextual encodings for word $i$ in the query $\mathcal{Q}$ and the document $\mathcal{D}$ respectively.

\subsection{Iterative Alternating Attention}
\label{sec:iterative}
This phase can be considered a means to uncover a possible inference chain that starts at the query and the document and leads to the answer.
The inference is modelled by an additional recurrent GRU network. The recurrent network iteratively performs an alternating search step to gather information that may be useful to predict the answer.
In particular, at each time step: (1) it performs an attentive read on the query encodings, resulting in a query glimpse, $\mathbf{q}_t$, and (2) given the current query glimpse, it extracts a conditional document glimpse, $\mathbf{d}_t$, representing the parts of the document that are relevant to the current query glimpse. In turn, both attentive reads are conditioned on the previous hidden state of the inference GRU $\mathbf{s}_{t-1}$, summarizing the information that has been gathered from the query and the document up to time $t$. The inference GRU uses both glimpses to update its recurrent state and thus decides which information needs to be gathered to complete the inference process.

\paragraph{Query Attentive Read} Given the query encodings $\{\mathbf{\tilde{q}}_i\}$, we formulate a query glimpse $\mathbf{q}_t$ at timestep $t$ by:
\[
q_{i,\, t} = \underset{i=1,\ldots,\mathcal{|Q|}}{\text{softmax}}\; \mathbf{\tilde{q}}_i^\top(\mathbf{A}_q \, \mathbf{s}_{t-1} + \mathbf{a}_q),
\]
\[
\mathbf{q}_t = \sum_i q_{i,\, t}\; \mathbf{\tilde{q}}_i
\]
where $q_{i,\, t}$ are the query attention weights and $\mathbf{A}_q \in \mathbb{R}^{2h \times s}$, where $s$ is the dimensionality of the inference GRU state, and $\mathbf{a}_q \in \mathbb{R}^{2h}$.
The attention we use here is similar to the formulation used in~\cite{hill2015goldilocks,sukhbaatar2015end}, but with two differences. First, we use a bilinear term instead of a simple dot product in order to compute the importance of each query term in the current time step. This simple bilinear attention has been successfully used in~\cite{luong-pham-manning:2015:EMNLP}. Second, we add a term $\mathbf{a}_q$ that allows to bias the attention mechanism towards words which tend to be important across the questions independently of the search key $\mathbf{s}_{t-1}$. 
This is similar to what is achieved by the original attention mechanism proposed in~\cite{bahdanau2014neural} without the burden of the additional $\tanh$ layer.

\paragraph{Document Attentive Read} The alternating attention continues by probing the document given the current query glimpse $\mathbf{q}_t$.
In particular, the document attention weights are computed based on both the previous search state and the currently selected query glimpse $\mathbf{q}_t$:
\[
d_{i,\, t} = \underset{i = 1, \ldots, \mathcal{|D|}}{\text{softmax}}\; \mathbf{\tilde{d}}_i^\top(\mathbf{A}_d \, [\mathbf{s}_{t-1}, \mathbf{q}_t] + \mathbf{a}_d),
\]
\[
\mathbf{d}_t = \sum_i d_{i,\, t}\; \mathbf{\tilde{d}}_i,
\]
where $d_{i,\, t}$ are the attention weights for each word in the document and $\mathbf{A}_d \in \mathbb{R}^{2h \times (s + 2h)}$ and $\mathbf{a}_d \in \mathbb{R}^{2h}$.
Note that the document attention is also conditioned on $\mathbf{s}_{t - 1}$. This allows the model to perform transitive reasoning on the document side,~i.e.~to use previously obtained document information to bias future attended locations, which is particularly important for natural language inference tasks~\cite{sukhbaatar2015end}.

\paragraph{Gating Search Results} In order to update its recurrent state, the inference GRU may evolve on the basis of the information gathered from the current inference step,~i.e. $\mathbf{s}_t = f([\mathbf{q}_t, \mathbf{d}_t], \mathbf{s}_{t-1})$, where $f$ is defined in Eq.~\ref{eq:gru}. However, the current query glimpse may be too general or the document may not contain the information specified in the query glimpse,~i.e.~the query or the document attention weights may be nearly uniform. We include a gating mechanism that is designed to reset the current query and document glimpses in the case that the current search is not fruitful. Formally, we implement a gating mechanism $\mathbf{r} = g([\mathbf{s}_{t-1}, \mathbf{q}_t, \mathbf{d}_t, \mathbf{q}_t \cdot \mathbf{d}_t])$, where $\cdot$ is the element-wise multiplication and $g : \mathbb{R}^{s + 6h} \rightarrow \mathbb{R}^{2h}$. The gate $g$ takes the form of a 2-layer feed-forward network with sigmoid output unit activation. The fourth argument of the gate takes into account multiplicative interactions between query and document glimpses, making it easier to determine the degree of matching between them. Given a query gate $g_q$, producing $\mathbf{r}_q$, and a document gate $g_d$, producing $\mathbf{r}_d$, the inputs of the inference GRU are given by the reset version of the query and document glimpses,~i.e., $\mathbf{s}_t = f([\mathbf{r}_q \cdot \mathbf{q}_t, \mathbf{r}_d \cdot \mathbf{d}_t], \mathbf{s}_{t-1})$. Intuitively, the model reviews the query glimpse with respect to the contents of the document glimpse and {\it vice versa}.

\subsection{Answer Prediction}
After a fixed number of time-steps $T$, the document attention weights obtained in the last search step $d_{i, T}$ are used to predict the probability of the answer given the document and the query $P(a | \mathcal{Q}, \mathcal{D})$. Formally, we follow~\cite{watson} and apply the ``pointer-sum'' loss:
\begin{equation}
\label{eq:answer_prob}
P(a | \mathcal{Q}, \mathcal{D}) = \sum_{i \in I(a, \mathcal{D})} d_{i, T},
\end{equation}
where $I(a, \mathcal{D})$ is a set of positions where $a$ occurs in the document. The model is trained to maximize $\log P(a | \mathcal{Q}, \mathcal{D})$ over the training corpus.

\section{Training Details}
\begin{table*}[t]
\small
\def\arraystretch{1.1}
\centering
\begin{tabular}{llccccc}
\toprule
\# & Model & \multicolumn{2}{c}{CBT-NE} & \phantom{ab} & \multicolumn{2}{c}{CBT-CN} \\
&& Valid & Test & & Valid & Test \\
\midrule
\small{1}&Humans (query) $^{1}$   &  - & 52.0 &  & - & 64.4 \\
\small 2&Humans (context+query) $^{1}$   & - & 81.6 &  & - & 81.6 \\
\midrule
\small3&LSTMs (context+query) $^{1}$    & 51.2  & 41.8 &  & 62.6 & 56.0 \\
\midrule
\small4&MemNNs (window memory + self-sup.) $^{1}$   & 70.4 & 66.6 &  & 64.2 & 63.0 \\
\small5&AS Reader $^{2}$                            & 73.8 & 
\textbf{68.6} &  & 68.8 & 63.4 \\
\midrule
\small6&Ours (fixed query attention) & 73.3 & 66.0 &  & 69.9 & 64.3 \\
\small7&Ours & \textbf{75.2} & \textbf{68.6} &  & \textbf{72.1} & \textbf{69.2} \\
\midrule
\midrule
\small8&AS Reader (Ensemble) $^2$ & 74.5 & 70.6 && 71.1 & 68.9 \\
\small9&Ours (Ensemble) & \textbf{76.9} & \textbf{72.0} && \textbf{74.1} & \textbf{71.0} \\
\bottomrule
\end{tabular}
\caption{\label{tab:cbt}Results on the CBT-NE (named entity) and CBT-CN (common noun) datasets. Results marked with $^1$ are from \protect\cite{hill2015goldilocks} and those marked with $^2$ are from \protect\cite{watson}.}
\end{table*}

To train our model, we used stochastic gradient descent with the ADAM optimizer~\cite{adam}, with an initial learning rate of 0.001. We set the batch size to 32 and we decay the learning rate by 0.8 if the accuracy on the validation set does not increase after a half-epoch,~i.e.~2000 batches (for CBT) and 5000 batches for (CNN).
We initialize all weights of our model by sampling from the normal distribution $\mathcal{N}$(0, 0.05). Following~\cite{saxe2013exact}, the GRU recurrent weights are initialized to be orthogonal and biases are initialized to zero.
In order to stabilize the learning, we clip the gradients if their norm is greater than 5~\cite{Pascanu13}.
We performed a hyperparameter search with embedding regularization in $\{0.001, 0.0001\}$, inference steps $T \in \{3, 5, 8\}$, embedding size $d \in \{256, 384\}$, encoder size $h \in \{128, 256\}$ and the inference GRU size $s \in \{256, 512\}$. We regularize our model by applying a dropout~\cite{srivastava2014dropout} rate of 0.2 on the input embeddings, on the search gates and on the inputs to both the query and the document attention mechanisms. We found that setting embedding regularization to $0.0001$, $T=8$, $d=384$, $h=128$, $s=512$ worked robustly across the datasets. Our model is implemented in Theano~\cite{Theano12}, using the Keras~\cite{chollet2015keras} library.

\paragraph{Computational Complexity} Similar to previous state-of-the-art models~\cite{watson,danqi} which use a bidirectional encoder, the major bottleneck of our method is computing the document and query encodings. The alternating attention mechanism runs only for a fixed number of steps ($T = 8$ in our tests), which is orders of magnitude smaller than a typical document or query in our datasets (see Table~\ref{tab:stats}).
The repeated attentions each require a softmax over ${\sim}$1000 locations which is typically fast on recent GPU architectures. Thus, our computation cost is comparable to~\cite{watson,danqi}, but we outperform the latter models on the datasets tested.

\section{Results}
We report the results of our model on the CBT-CN, CBT-NE and CNN datasets, previously described in Section~\ref{sec:task}.


\begin{table*}[t]
\small
\def\arraystretch{1.1}
\centering
\begin{tabular}{llccc}
\toprule
\# & Model & \phantom{abcd} & \multicolumn{2}{c}{CNN} \\
&& & Valid & Test \\
\midrule
\small{1}&Word distance model $^{1}$ && 50.5 & 50.9 \\
\small{2}&Deep LSTM Reader $^{1}$  && 55.0 & 57.0 \\
\small{3}&Attentive Reader $^{1}$  && 61.6 & 63.0 \\
\small{4}&Impatient Reader $^{1}$                              && 61.8 & 63.8 \\
\midrule
\small{5}&MemNNs (window memory) $^{2}$                        && 50.8 & 60.6 \\
\small{6}&MemNNs (window memory + self sup.) $^{2}$            && 63.4 & 66.8 \\
\small{7}&AS Reader $^{3}$                                     && 68.6 & 69.9 \\
\small{8}&Ours                                                 && \textbf{72.6} & \textbf{73.3} \\
\midrule
\small{9}&Stanford AR (with GloVe) $^{4}$                      && 72.4 & 72.4 \\
\midrule
\midrule
\small{10}&MemNNs (Ensemble) $^{2}$                            && 66.2 & 69.4 \\
\small{11}&AS Reader (Ensemble) $^3$                           && 73.9 & 75.4 \\
\small{13}&Ours (Ensemble)                                     && \textbf{75.2} & \textbf{76.1} \\
\bottomrule
\end{tabular}
\caption{\label{tab:cnn}Results on the CNN datasets. Results marked with $^1$ are from \protect\cite{hermann2015teaching}, $^2$ from \protect\cite{hill2015goldilocks}, $^{3}$ from \protect\cite{watson} and $^4$ from~\protect\cite{danqi}.}
\end{table*}

\subsection{CBT}
Table~\ref{tab:cbt} reports our results on the CBT-CN and CBT-NE dataset. The Humans, LSTMs and Memory Networks (MemNNs) results are taken from~\cite{hill2015goldilocks} and the Attention-Sum Reader (AS Reader) is a state-of-the-art result recently obtained by~\cite{watson}.

\paragraph{Main result} Our model (line 7) sets a new state-of-the-art on the common noun category by gaining 3.6 and 5.6 points in validation and test over the best baseline AS Reader (line 5). This performance gap is only partially reflected on the CBT-NE dataset. We observe that the 1.4 accuracy points on the validation set do not reflect better performance on the test set, which sits on par with the best baseline. In CBT-NE, the missing word is a named entity appearing in the story which is likely to be less frequent than a common noun. We found that approximatively 27.5\% of validation examples and 29.6\% of test examples contain an answer that has never been predicted in the training set. These numbers are considerably lower for the CBT-CN, for which only 2.5\% and 4.6\% of validation and test examples respectively contain an answer that has not been previously seen.

\paragraph{Ensembles} Fusing multiple models generally achieves better generalization.
In order to investigate whether this could help achieving better held-out performance on CBT-NE, we adopt a simple strategy and average the predictions of 5 models trained with different random seeds (line 9). In this case, our ensemble outperforms the AS Reader ensemble both on CBT-CN and CBT-NE setting new state-of-the-art for this task. On CBT-NE, it achieves a validation and test performance of 76.9 and 72.0 accuracy points respectively (line 9). On CBT-CN it shows additional improvements over the single model and sits at 74.1 on validation and 71.0 on test.

\paragraph{Fixed query attention} In order to measure the impact of the query attention step in our model, we constrain the query attention weights $q_{i, t}$ to be uniform,~i.e. $q_{i, t} = 1/|\mathcal{Q}|$, for all $t = 1,\ldots,T$ (line 6). This corresponds to fixing the query representation to the average pooling over the bidirectional query encodings and is similar in spirit to previous work~\cite{watson,danqi}. By comparing line 6 and line 7, we see that the query attention mechanism allows improvements up to 2.3 points in validation and 4.9 points in test with respect to fixing the query representation throughout the search process. A similar scenario was observed on the CNN dataset.

\subsection{CNN}
Table~\ref{tab:cnn} reports our results on the CNN dataset. We compare our model with a simple word distance model, the three neural approaches from~\cite{hermann2015teaching} (Deep LSTM Reader, Attentive Reader and Impatient Reader), and with the AS reader~\cite{watson}.

\paragraph{Main result} The results show that our model (line 8) improves state-of-the-art accuracy by 4 percent absolute on validation and 3.4 on test with respect to the most recent published result (AS Reader) (line 7). We also report the very recent results of the Stanford AR system that came to our attention during the write-up of this article~\cite{danqi} (line 9).
Our model slightly improves over this strong baseline by 0.2 percent on validation and 0.9 percent on test. We note that the latter comparison may be influenced by different training and initialization strategies. First, Stanford AS uses GloVe embeddings~\cite{pennington2014glove}, pre-trained from a large external corpus.
Second, the system normalizes the output probabilities only over the candidate answers in the document.

\paragraph{Ensembles} We also report the results using ensembled models. Similarly to the single model case, our ensembles achieve state-of-the-art test performance of 75.2 and 76.1 on validation and test respectively, outperforming previously published results.

\begin{table}[t]
    \centering
    \vspace{2mm}
    \begin{tabular}{llrlr}
    \toprule
    Category & \multicolumn{2}{c}{Stanford AR} & \multicolumn{2}{c}{Ours} \\
    \midrule
    
    Exact Match        & 13 & (100.0\%) & 13 & (100.0\%) \\
    Paraphrasing       & 39 & (95.1\%)  & 39 & (95.1\%)  \\
    Partial Clue        & 17 & (89.5\%)  & 16 & (84.2\%)  \\
    \midrule
    Multiple Sent.     & 1  & (50.0\%)  & 1  & (50.0\%)  \\
    Ambig. / Hard   & 1  & (5.9\%)   & 5  & (29.4\%)   \\
    \midrule
    Coref. Errors & 3  & (37.5\%)  & 3  & (37.5\%)  \\
    \midrule
    All                & 74 &   & 77 &      \\
    \bottomrule
    \end{tabular}
    \label{tab:cnn-category}
    \caption{Per-category performance of the Stanford AR and our system. The first three categories require local context matching, the next two global context matching and coreference errors are unanswerable questions~\protect\cite{danqi}.}
\end{table}

\paragraph{Category analysis}~\cite{danqi} classified a sample of 100 CNN stories based on the type of inference required to guess the answer. Categories that only require local context matching around the placeholder and the answer in the text are Exact Match, Paraphrasing, and Partial Clue, while those which require higher reasoning skills are Multiple Sentences and Ambiguous. For example, in Exact Match examples, the question placeholder and the answer in the document share several neighboring exact words.

Category-specific results are reported in Table~\ref{tab:cnn-category}. Local context categories generally seem to be easily tackled by the neural models, which perform similarly. It seems that the iterative alternating attention inference is better able to solve more difficult examples such as Ambiguous/Hard. One hypothesis is that, in contrast to Stanford AR, which uses only one fixed-query attention step, our iterative attention may better explore the documents and queries. Finally, Coreference Errors ($\sim$25\% of the corpus) includes examples with critical coreference resolution errors which may make the questions ``unanswerable''. This is a barrier to achieving accuracies considerably above 75\%~\cite{danqi}. If this estimate is accurate, our ensemble model (76.1\%) may be approaching near-optimal performance on this dataset.


\begin{figure}
    \centering            
    \includegraphics[width=\linewidth]{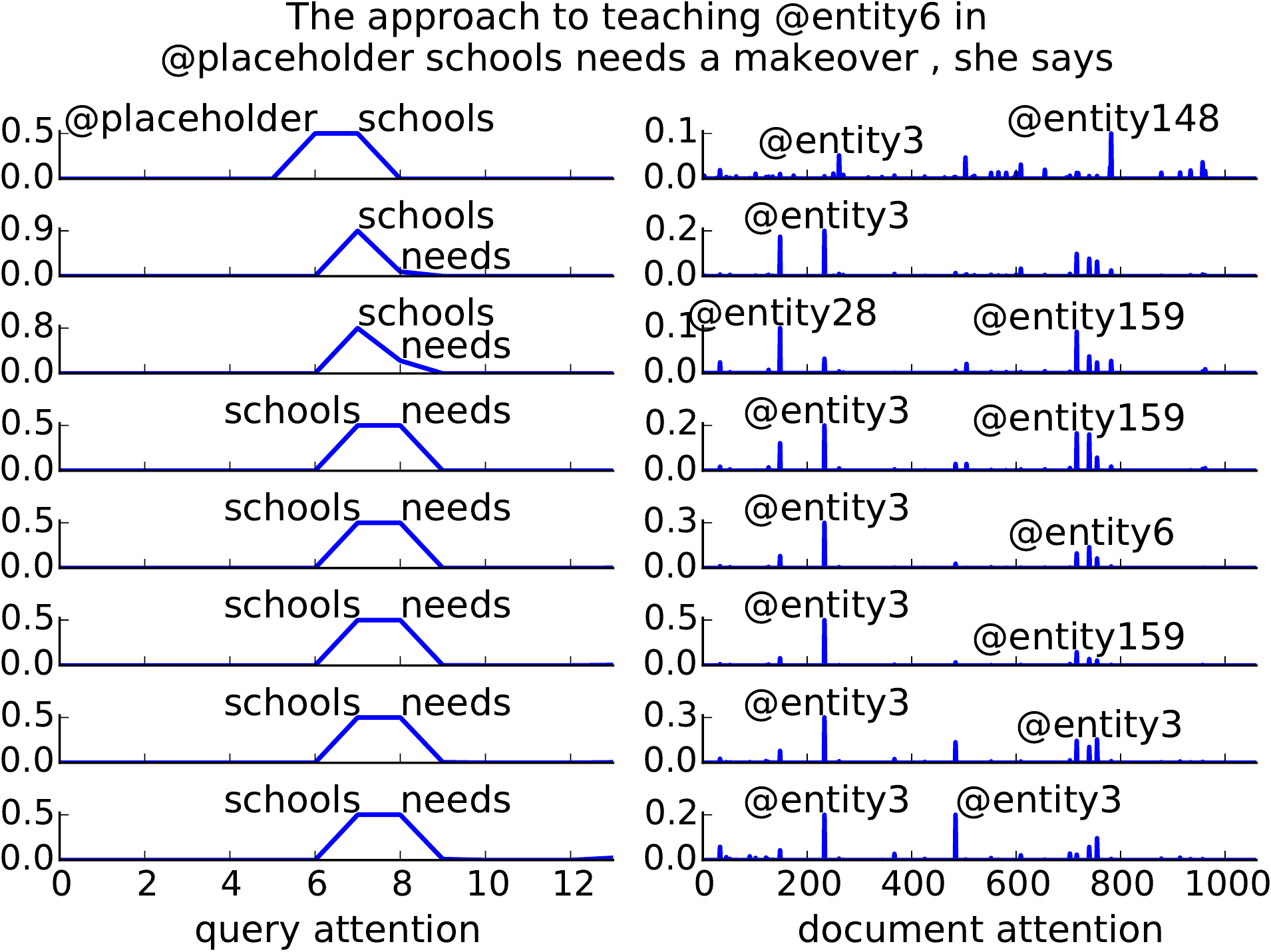}
    \label{fig:attention-plot}
    \vspace{-3mm}
    \caption{Visualization of the alternated attention mechanism for an article in CNN, treating about the decline of the Italian language in schools. The title of the plot is the query. Each row correspond to a timestep. The target is @entity3 which corresponds to the word ``Italian''.}
\end{figure}

\subsection{Discussion}

We inspect the query and document attention weights for an example article from the CNN dataset. The title of the article is ``Dante turns in his grave as Italian language declines'', and it discusses the decline of Italian language in schools. The plot is shown in Figure~\ref{fig:attention-plot}, where locations attended to in the query and document are in the left and right column respectively. Each row corresponds to an inference timestep $1 \leq t \leq 8$. At the first step, the query attention focuses on the placeholder token, as its local context is generally important to discriminate the answer. The model first focuses on @entity148, which corresponds to ``Greek'' in this article. At this point, the model is still uncertain about other possible locations in the document (we can observe small weights across document locations). At $t = 2$, the query attention moves towards ``schools'' and the model hesitates between ``Italian'' and ``European Union'' (@entity28, see step 3), both of which may satisfy the query.
At step 3, the most likely candidates are ``European Union'' and ``Rome'' (@entity159). As the timesteps unfold, the model learns that ``needs'' may be important to infer the correct entity,~i.e. ``Italian''. The query sits on the same attended location, while the document attention evolves to become more confident about the answer.

We find that, across CBT and CNN examples, the query attention wanders near or focuses on the placeholder location, attempting to discriminate its identity using only local context. For these particular datasets, the majority of questions can be answered after attending only to the words directly neighbouring the placeholder. This aligns with the findings of~\cite{danqi} concerning CNN, which state that the required reasoning and inference levels for this dataset are quite simple. It would be worthwhile to formulate a dataset in which the placeholder is harder to infer using only local neighboring words, and thereby necessitates deeper query exploration.

Finally, across this work we fixed the number of inference steps $T$. We found that using $8$ timesteps works well consistently across the tested datasets. However, we hypothesize that more (fewer) time-steps would benefit harder (easier) examples. A straight-forward extension of the model would be to dynamically select the number of inference steps conditioned on each example.

\section{Related Works}

Neural attention models have been applied recently to a sm\"org\aa sbord of machine learning and natural language processing problems. These include, but are not limited to, handwriting recognition~\cite{Graves13}, digit classification~\cite{mnih2014recurrent}, machine translation~\cite{bahdanau2014neural}, question answering~\cite{sukhbaatar2015end,hermann2015teaching} and caption generation~\cite{xu2015show}.
In general, attention models keep a memory of states that can be accessed at will by learned attention policies.
In our case, the memory is represented by the set of document and query contextual encodings.

Our model is closely related to~\cite{sukhbaatar2015end,kumar2015ask,hermann2015teaching,watson,hill2015goldilocks}, which were also applied to question answering.
The pointer-style attention mechanism that we use to perform the final answer prediction has been proposed by~\cite{watson}, which in turn was based on the earlier Pointer Networks of~\cite{vinyals2015pointer}.  However, differently from our work,~\cite{watson} perform only one attention step and embed the query into a single vector representation, corresponding to the concatenation of the last state of the forward and backward GRU networks. To our knowledge, embedding the query into a single vector representation is a choice that is shared by most machine reading comprehension models. In our model, the repeated, tight integration between query attention and document attention allows the model to explore dynamically which parts of the query are most important to predict the answer, and then to focus on the parts of the document that are most salient to the currently-attended query components.

A similar attempt in attending different components of the query may be found in~\cite{hermann2015teaching}. In that model, the document is processed once for each query word. This can be computationally intractable for large documents, since it involves unrolling a bidirectional recurrent neural network over the entire document multiple times. In contrast, our model only estimates query and document encodings once and can learn how to attend different parts of those encodings in a fixed number of steps. The inference network is responsible for making sense of the current attention step with respect to what has been gathered before. In addition to achieving state-of-the-art performance, this technique may also prove to be more scalable than alternative query attention models.

Finally, our iterative inference process shares similarities to the iterative \emph{hops} in Memory Networks~\cite{sukhbaatar2015end,hill2015goldilocks}. In that model, the query representation is updated iteratively from hop to hop, although its different components are not attended to separately. Moreover, we substitute the simple linear update with a GRU network. The gating mechanism of the GRU network made it possible to use multiple steps of attention and to propagate the learning signal effectively back through to the first timestep.

\section{Conclusion}
We presented an iterative neural attention model and applied it to machine comprehension tasks. Our architecture deploys a novel alternating attention mechanism, and tightly integrates successful ideas from past works in machine reading comprehension to obtain state-of-the-art results on three datasets. The iterative alternating attention mechanism continually refines its view of the query and document while aggregating the information required to answer a query.

Multiple future research directions may be envisioned. We plan to dynamically select the optimal number of inference steps required for each example. Moreover, we suspect that shifting towards stochastic attention should permit us to learn more interesting search policies. Finally, we believe that our model is fully general and may be applied in a straightforward way to other tasks such as information retrieval.

\bibliographystyle{emnlp2016}
\bibliography{twoway}

\end{document}